\documentclass[10pt,twocolumn,letterpaper]{article}

\usepackage[pagenumbers]{wacv} 
\usepackage{caption}
\usepackage{graphicx}
\usepackage{cuted}

\newcommand{\boldheader}[1]{\textbf{#1}}

\definecolor{wacvblue}{rgb}{0.21,0.49,0.74}
\usepackage[pagebackref,breaklinks,colorlinks,allcolors=wacvblue]{hyperref}

\def\wacvPaperID{1037} 
\def\confName{WACV}
\def\confYear{2027}

\title{Does Video Memory Use What It Retrieves? \\ A Causal Audit of Memory Specificity}

\author{
Aditi Tiwari$^{1,2}$\thanks{Work done during internship at Adobe Research.}
\quad
Akshit Bhalla$^{2}$
\quad
Darshan Prasad$^{2}$
\quad
Heng Ji$^{1}$
\\[3pt]
$^{1}$University of Illinois Urbana-Champaign
\qquad
$^{2}$Adobe Research
\\[2pt]
{\tt\small
\{aditit5,hengji\}@illinois.edu
\qquad
\{akshitb,dprasad\}@adobe.com
}
}

\begin{document}
\maketitle

\begin{abstract}
Video models increasingly use memory to preserve information over long sequences, with the assumption that gains come from retrieving and using the correct past content. Standard memory ablations test whether memory helps, but not whether the retrieved content is responsible. We test this directly with \emph{read-time memory substitution}, which replaces the consumed memory value while leaving the rest of the computation unchanged. This separates \emph{memory benefit} from \emph{memory specificity}, the extent to which the gain depends on retrieved content. Across frozen video world models, identity-free controls containing no evaluation-specific content recover essentially the full benefit on Ego-Exo4D and 7-Scenes and about $70\%$ on TUM. In the Ego-Exo4D dose response, recovery falls from $102\%$ to $1\%$ as these values move away from observed training-memory representations, supporting representation repair as the best-supported explanation in this setting. WorldMem shows graded dependence. A wrong memory from the same trajectory recovers $94.1\%$ of the PSNR benefit relative to zero content, while a donor from a disjoint trajectory and biome recovers $43.7\%$. SAM 2 shows strong content dependence. On DAVIS, replacing the correct spatial memory with a valid wrong memory reduces mean region and boundary score from $0.926$ to $0.182$. At MOSEv2 reappearance, it falls from $0.459$ to $0.000$. These results show that memory gains can depend on generic representation support, broader context, or exact episodic content. Read-time substitution provides a direct way to distinguish them.
\end{abstract}
\vspace{-1em}   
\section{Introduction}
\label{sec:intro}

Long-horizon video models often need information that appeared much earlier in a sequence. World models can drift from previously observed scene states during long rollouts \cite{wu2025spatialmemory,xiao2025worldmem}. Video segmentation models can lose a target during occlusion and must recover it when the object returns \cite{xmem,sam2,sam2long}. Memory is a natural solution. The model stores information from earlier frames and retrieves it when current evidence is insufficient.

Many memory systems rely on an implicit assumption. Their benefit should depend on retrieving the correct past content \cite{xiao2025worldmem,worldpack,xmem,sam2}. If a model needs information from a particular earlier event, the correct memory should help more than an unrelated one. This assumption motivates better retrieval, larger memory stores, and more selective memory access \cite{oh2019videoobjectsegmentationusing}.

Yet standard memory evaluations rarely test this assumption directly. They typically compare performance with memory against performance without memory \cite{xiao2025worldmem,worldpack,xmem,sam2,sam2long}. This measures how much memory helps, which we call the \emph{memory benefit}. It does not measure how much of that gain depends on the information that was retrieved. An unrelated but plausible memory could produce a similar improvement without supplying the correct past content.

We therefore distinguish memory benefit from \emph{memory specificity}. Memory specificity measures how much of the memory benefit depends on information carried by the retrieved content. A system can have a large memory benefit but low specificity if an unrelated or identity-free memory works almost as well as the correct one. One possible explanation for such identity-independent benefit is \emph{representation repair}, where memory improves the current representation without supplying episode-specific information. More generally, memory dependence can occur at different levels. A model may require only generic representation support, broader scene or trajectory context, or the exact past event.

This distinction matters for memory design. Better retrieval can only improve the part of the memory gain that depends on what was retrieved. If the same gain survives with identity-free memory, retrieval is not the active ingredient in that part of the improvement. If broader context matters but exact identity does not, retrieval may need to preserve the relevant context rather than recover the exact past state. This leads to a simple question. \emph{When memory improves a video model, how much of that improvement requires the retrieved content to be correct?}

We answer this question with \emph{read-time memory substitution}. At the moment a model consumes memory, we replace the retrieved value while keeping the rest of the computation fixed. We compare the correct memory with a valid memory from an unrelated scene or object, an identity-free value constructed without evaluation-specific information, and a value far from observed training-memory representations. We also compare wrong memories that preserve different amounts of scene or trajectory context. The resulting performance ladder reveals what information the memory must preserve for the gain to remain.

Across frozen video world models, most of the measured memory benefit survives without the correct past content. On Ego-Exo4D and 7-Scenes, identity-free memory recovers essentially the full benefit of correct memory. On TUM, it recovers about $70\%$, suggesting a mixed regime with a larger content-specific residual. On Ego-Exo4D, recovery falls as identity-free values move away from observed training-memory representations and also collapses when injection direction changes at fixed magnitude. These interventions support representation repair as the best-supported explanation in this setting.

We next test whether limited exact-content dependence is specific to our trained memory pathway. WorldMem \cite{xiao2025worldmem}, a generative world model with a native trained memory bank, shows graded content dependence. A wrong memory from the same trajectory recovers about $94\%$ of the PSNR benefit relative to zero content, while a memory from a disjoint trajectory and biome still recovers about $43\%$. Exact place identity therefore adds little beyond same-trajectory context, while broader contextual agreement remains important.

SAM 2 provides the contrasting case. Its native memory was trained as part of the video object segmentation system \cite{sam2}. On DAVIS \cite{davis}, replacing the correct spatial memory with a valid memory from another object reduces the mean region and boundary segmentation score from $0.926$ to $0.182$. An identity-free prototype gives $0.126$. The same dependence persists at object reappearance on MOSEv2 \cite{mose,mosev2}, where the score falls from $0.459$ with correct spatial memory to $0.000$ with wrong spatial memory and $0.003$ with an identity-free prototype. The same audit therefore detects strong content dependence when the prediction requires the correct stored information.


Our contributions are threefold.
\begin{itemize}

\item We distinguish \emph{memory benefit} from \emph{memory specificity}, separating how much memory helps from how much of that gain depends on retrieved content.
\item We introduce \emph{read-time memory substitution}, which measures content dependence while preserving the model, memory interface, and preceding trajectory.
\item We uncover different forms of memory use across model families. Frozen world models show limited content specificity, with Ego-Exo4D supporting representation repair as the best-supported explanation. WorldMem depends strongly on broader context but little on the exact place, while SAM 2 depends strongly on correct spatial memory.

\end{itemize}

\section{Related Work}
\label{sec-related}

\boldheader{External memory and retrieval augmentation.}
External-memory models introduced learned read and write operations, explicit addressing, and key-value access \cite{ntm,memorynet,dnc,keyvaluemem}. Retrieval-augmented models later scaled this idea to large non-parametric stores that condition prediction and generation \cite{knnlm,realm,rag,retro}. These methods show that retrieval can help, but not how much the gain depends on the retrieved content.

\boldheader{Memory in video.}
Video memory is used to maintain object identity, extend temporal context, and preserve scene state over long sequences \cite{oh2019videoobjectsegmentationusing,xmem,sam2,sam2long,memvit,malmm,xiao2025worldmem,worldpack}. Recent work also studies compressed history, alternative readout mechanisms, and state-space memory designs \cite{worldtrace,echomem,lcss}. Most of this work asks how memory should be stored or retrieved. We ask what information in the retrieved memory is responsible for the gain.

\boldheader{Retrieval, utilization, and causal interventions.}
Access to relevant information does not guarantee that a model uses it \cite{lostinthemiddle,instructretro}. Causal mediation, interchange interventions, and causal tracing distinguish information that is present from information that causally affects the output by modifying internal representations while preserving the surrounding computation \cite{vigcausal,geigerinterchange,mengrome}. We apply the same principle directly to the consumed memory value.

\boldheader{Identity-independent representation repair.}
Denoising, feature alignment, and test-time adaptation can improve representations without supplying instance-specific information \cite{vincentdae,alainbengio,deepcoral,tent}. This motivates an alternative explanation for memory gains, where memory improves the current representation without recovering the correct past event. Our audit separates this generic representation support from broader contextual dependence and exact retrieved content.

\begin{figure*}[h]
\centering
\includegraphics[width=\textwidth]{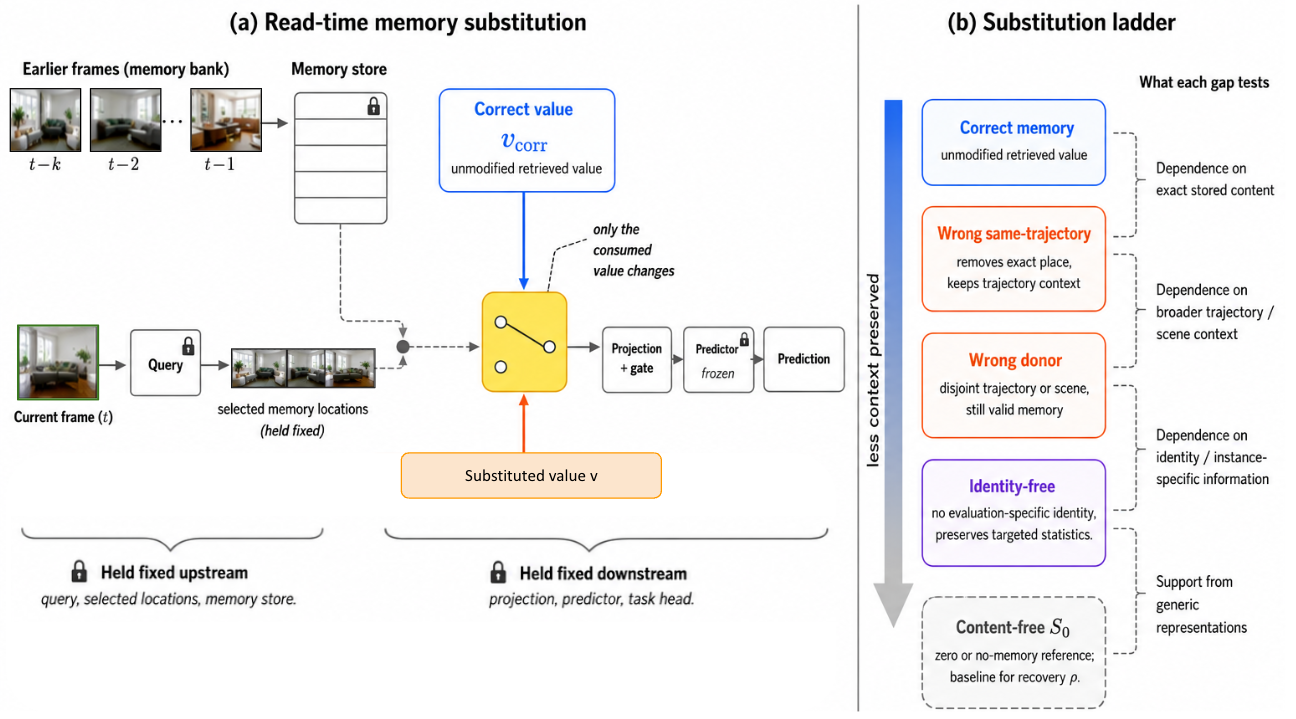}
\caption{Read-time memory substitution and the substitution ladder. (a) At a target memory read, only the consumed value is changed while the memory trajectory, query, selected locations, and downstream model remain fixed. (b) Substitutions progressively remove exact content, broader context, and identity-specific information. The content-free condition $S_0$ serves as the reference for benefit recovery.}
\label{fig-method-overview}
\end{figure*}


\section{A Causal Audit of Memory Specificity}
\label{sec-method}
We measure memory specificity by intervening on the value consumed at a memory read while preserving the surrounding computation. This separates dependence on memory content from the effect of having a memory pathway at all.

\subsection{Memory Benefit and Memory Specificity}
\label{sec-method-specificity}

Let $S(v)$ denote task performance when the model consumes memory value $v$, where larger $S$ is better, and let $v_{\mathrm{corr}}$ denote the correct memory. We call the intervention that removes memory content the \emph{content-free reference}, and let $S_0$ denote performance under this reference. When supported by the model, the reference is the no-memory condition. Otherwise, it uses zero memory content while preserving the native memory interface. We define the memory benefit in Eq.~\ref{eq-memory-benefit} as
{\small
\begin{equation}
B = S(v_{\mathrm{corr}}) - S_0.
\label{eq-memory-benefit}
\end{equation}
}
For models with a no-memory condition, this is the usual memory-on versus memory-off comparison.

We measure how much of this benefit survives under a substituted value $v$ using \emph{benefit recovery}, or recovery for short:
{\small
\begin{equation}
\rho(v) =
\frac{S(v)-S_0}
{S(v_{\mathrm{corr}})-S_0}.
\label{eq-benefit-recovery}
\end{equation}
}
In Eq.~\ref{eq-benefit-recovery}, $\rho(v)=1$ means full recovery and $\rho(v)=0$ means no recovery. The remaining fraction, $1-\rho(v)$, is the content-specific residual relative to that control. We report recovery only when correct memory provides a positive benefit. We do not clip $\rho(v)$, so values can fall slightly below zero or above one.

We assess \emph{memory specificity} from how recovery changes as information is removed. High recovery under an identity-free control means that little of the benefit requires evaluation-specific content. Low recovery means that the removed information matters. Because different controls remove different information, specificity is intervention-dependent. We therefore report recovery for each control rather than reduce the ladder to one score.

We use \emph{representation repair} for the hypothesis that identity-independent memory improves the current representation without supplying episode-specific information. Low specificity alone does not establish this mechanism. We test its additional predictions in Sec.~\ref{sec-method-repair}.

\subsection{Read-Time Memory Substitution}
\label{sec-method-substitution}

We isolate memory content with \emph{read-time memory substitution}. At a target step, we replace only the value consumed by the memory pathway while preserving the surrounding computation, as shown in Figure~\ref{fig-method-overview}. The query, selected memory locations, number of memory entries, temporal and positional information, network weights, and all earlier computation remain unchanged. This intervenes on memory content rather than removing the memory pathway \cite{pearl2009causality, geigerinterchange}.

The earlier memory trajectory is also preserved. The intervention therefore measures the direct effect of the consumed value at the target step. It does not measure effects from changing earlier memory reads.

The audit requires access to the memory-read interface but does not require retraining. We can therefore apply it to systems with different memory architectures and training procedures.

\subsection{The Substitution Ladder}
\label{sec-method-ladder}

We evaluate substitutions that remove different amounts of information while preserving the read interface. A single wrong-memory control is insufficient because it can change both exact identity and broader contextual agreement. Figure~\ref{fig-method-overview}(b) summarizes the substitution ladder. Each step removes additional information while preserving the memory interface.

The \emph{correct memory}, $v_{\mathrm{corr}}$, is used by the unmodified system. A \emph{valid wrong memory}, $v_{\mathrm{wrong}}$, comes from the same memory system but does not correspond to the current scene, place, or object. An \emph{identity-free memory}, $v_{\mathrm{idfree}}$, contains no evaluation-specific identity while preserving the statistics targeted by that control. Its construction depends on the memory family and is defined in Sec.~\ref{sec-setup}. Identity-free memory still preserves memory-like structure. The content-free reference instead removes the memory content used to define the benefit. Strongly perturbed values provide additional controls.

Wrong memories can also preserve different amounts of context. A wrong memory from the same trajectory removes the exact place while retaining broader trajectory and scene information. A donor from a disjoint trajectory removes the exact correspondence and much of this shared context. These controls probe dependence on generic representation support, broader context, and exact episodic content.

The recovery pattern determines the interpretation. Low recovery under wrong and identity-free memories indicates strong content dependence. High identity-free recovery means that exact episodic content contributes little relative to that control. Intermediate recovery indicates additional benefit from broader context or exact content.

\subsection{Testing the Representation-Repair Hypothesis}
\label{sec-method-repair}

Low memory specificity does not explain why an identity-free value helps. Representation repair makes two additional predictions. Recovery should decrease as the value moves away from memory representations observed during training. It should also depend on representation direction when injection magnitude is fixed.

Let $\mathcal{M}_{\mathrm{train}}$ be the training-memory read representations. We define distance to the nearest observed training memory in Eq.~\ref{eq-dreal}:
{\small
\begin{equation}
d_{\mathrm{real}}(v) =
\min_{m \in \mathcal{M}_{\mathrm{train}}} d(v,m),
\label{eq-dreal}
\end{equation}
}
where $d(v,m)=1-\frac{v^\top m}{\lVert v\rVert\,\lVert m\rVert}$ is cosine distance in the model's memory-read representation space. The reference set contains training memories only, so it contains no values from the evaluation episode. Small $d_{\mathrm{real}}$ means that the injected value lies near an observed training memory.

We test the first prediction with a dose response. Starting from an identity-free training-memory value, we progressively perturb it away from observed training memories and measure $d_{\mathrm{real}}$ and $\rho(v)$. Representation repair predicts lower recovery as $d_{\mathrm{real}}$ increases.

We test the second prediction by holding injection magnitude fixed while changing direction away from a high-recovery direction. If magnitude alone explains the gain, equal-magnitude interventions should behave similarly. A drop in recovery shows that injection strength alone is insufficient.

\subsection{Family-Specific Readouts}
\label{sec-method-readouts}

The intervention is shared across model families, but the task metric differs.

For latent world models, we measure \emph{revisit consistency error} (RCE), the cosine distance between the predicted revisit latent and the target latent from the earlier observation. Lower RCE is better. Recovery is therefore
{\small
\begin{equation}
\rho_{\mathrm{RCE}}(v) =
\frac{\mathrm{RCE}_0-\mathrm{RCE}_v}
{\mathrm{RCE}_0-\mathrm{RCE}_{\mathrm{corr}}},
\label{eq-rce-recovery}
\end{equation}
}
where $\mathrm{RCE}_0$ is the error under the content-free reference and $\mathrm{RCE}_{\mathrm{corr}}$ is the error with correct memory. Eq.~\ref{eq-rce-recovery} gives one for full recovery and zero for no improvement over the reference.

For generative world models, we measure reconstruction quality beyond the context window using Peak Signal-to-Noise Ratio (PSNR) and Learned Perceptual Image Patch Similarity (LPIPS). Higher PSNR and lower LPIPS are better. PSNR follows Eq.~\ref{eq-benefit-recovery}. For LPIPS, we use Eq.~\ref{eq-rce-recovery} with LPIPS in place of RCE.

For video object segmentation, we use region similarity $\mathcal{J}$ and boundary accuracy $\mathcal{F}$ and report their mean, $\mathcal{J}\&\mathcal{F}$ \cite{davis, Perazzi2016ABD}. Higher values are better. We measure the target frame under each substitution and recovery over later frames for reappearance experiments. Target transfer records whether the prediction overlaps another annotated object more strongly than the intended target.

\begin{figure*}[t]
\centering
\includegraphics[width=\textwidth]{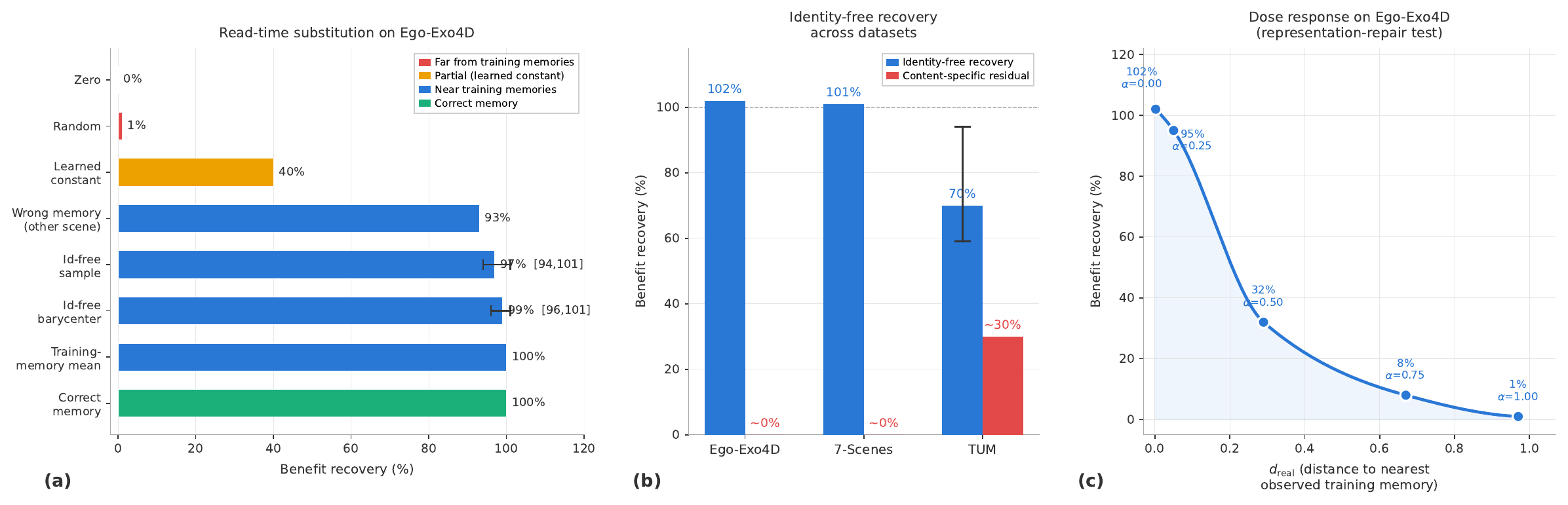}
\caption{World-model memory specificity and mechanism on DINO-WM.
(a) Read-time substitution on Ego-Exo4D. Identity-free and wrong
memories retain nearly the full memory benefit, while zero and random
controls do not. A learned constant provides only partial recovery.
(b) Training-memory mean recovery across datasets using controls
constructed only from training memories. Recovery is near complete on
Ego-Exo4D and 7-Scenes, while TUM shows a less certain mixed regime.
(c) Dose response on Ego-Exo4D. An identity-free value constructed
from training memories is progressively perturbed away from observed
training-memory representations. Benefit recovery decreases as
$d_{\mathrm{real}}$ increases.}
\label{fig-wm-main}
\end{figure*}


\section{Experimental Setup}
\label{sec-setup}

We apply read-time memory substitution to three video memory settings. These are a trained memory pathway attached to frozen video world models, the native memory of WorldMem, and the native memory of SAM 2. The intervention is shared across settings, while the memory representation and task readout differ.

\subsection{Frozen Video World Models}
\label{sec-setup-worldmodels}

Our main experiments use DINO-WM \cite{dinowm}, which predicts future states in the feature space of a frozen DINOv2 encoder \cite{dinov2}. We freeze the encoder and DINO-WM predictor and train only a small memory pathway. The memory stores pose-keyed latent states, retrieves a bounded set of past states, and injects the resulting read through a learned projection and gate. We use DINO-WM for the full substitution ladder, dose response, and fixed-magnitude controls. We also attach the same memory design to an action-conditioned V-JEPA 2 host \cite{vjepa2} and use the correct-memory versus wrong-memory contrast as a second-host check.

\boldheader{Datasets.}
We evaluate on TUM RGB-D \cite{tumrgbd}, Ego-Exo4D \cite{egoexo4d}, and 7-Scenes \cite{sevenscenes}. Each dataset contains trajectories with repeated observations of locations or environments. We identify revisit events and compare the predicted latent at a revisit with the latent associated with the corresponding earlier observation. For 7-Scenes, memory construction and evaluation use disjoint scenes.

\boldheader{Memory controls.}
We call a control \emph{leakage-safe} when it is constructed only from training memories and never uses the evaluation trajectory. All training-memory controls below satisfy this condition. We evaluate the mean of training-memory reads, a barycenter in memory-read space, and an identity-free sample in read space. Random and strongly perturbed values provide controls far from observed training-memory representations. We also evaluate a \emph{learned constant}, a single vector $c\in\mathbb{R}^{d}$ optimized on training revisit pairs while the encoder, predictor, projection, and gate remain frozen. The same $c$ is injected for every evaluation query, with no retrieval or query-specific information. This control tests how much of the memory benefit can be reproduced by one fixed input to the memory pathway. The same training reads define $\mathcal{M}_{\mathrm{train}}$ in Eq.~\ref{eq-dreal}. For the dose response, we perturb an identity-free value constructed from training memories away from observed training-memory representations. For the fixed-magnitude control, we hold the injected norm constant while changing its direction.

\subsection{WorldMem}
\label{sec-setup-worldmem}

We use WorldMem \cite{xiao2025worldmem}, a generative world model for Minecraft with a native memory bank of past frames. WorldMem uses a Conditional Diffusion Transformer with Diffusion Forcing and retrieves past frames using pose, field-of-view overlap, and recency. We use the official pretrained checkpoint and keep all model parameters frozen.

\boldheader{Data and evaluation.}
We evaluate $1000$ cases from the WorldMem Minecraft test split, covering plains, desert, iceplains, and savanna biomes. Following the WorldMem beyond-context-window setting, the model receives $600$ context frames and generates beyond this context within a $700$-frame sequence. We report PSNR and LPIPS on the generated region as defined in Sec.~\ref{sec-method-readouts}.

\boldheader{Memory controls.}
The \emph{correct} condition uses the memory selected by the unmodified model. The \emph{wrong-place} condition substitutes a real memory from another place in the same trajectory. The \emph{wrong-donor} condition uses a real memory from a disjoint trajectory and different biome, removing exact place correspondence and much of the shared context. The \emph{mean-content} condition uses the mean retrieved content from the same scene, removing exact frame identity while retaining broader scene statistics. The \emph{zero-content} condition replaces the retrieved content with zeros. Because WorldMem does not support removing its memory interface in this evaluation, zero content serves as the content-free reference $S_0$. The wrong-place and wrong-donor controls preserve different amounts of context and therefore probe dependence on exact place versus broader trajectory and scene information.

\subsection{SAM 2}
\label{sec-setup-sam2}

We use SAM 2 with the Hiera small backbone and its native video memory \cite{sam2}. The pretrained model is left unchanged. SAM 2 conditions each target frame on spatial mask-conditioned memory features and an object pointer. We replace the spatial memory with memory from another object, replace the object pointer with one from another object, or replace both. We also evaluate an identity-free prototype formed from donor memories and a zero-memory control. Donors come from disjoint videos.

\boldheader{DAVIS 2017.}
We first evaluate on DAVIS 2017 \cite{davis}. We apply the substitution at a fixed target point in each sequence and measure segmentation quality at that frame. This evaluates dependence on the spatial memory and object pointer when the target is generally visible.

\boldheader{MOSEv2 reappearance.}
We evaluate disappearance and reappearance on MOSEv2 \cite{mosev2,mose}. We use the training split because dense frame-level masks are needed to identify reappearance events and score the returning target. This is an analysis of memory use rather than a leaderboard evaluation. A reappearance event contains a target that is visible, absent for at least five frames, and then visible again. SAM 2 is initialized from the first annotated target frame and propagated through the disappearance. We substitute the memory only at the first reappearance frame, leaving the preceding memory trajectory unchanged.

At reappearance, we report mean region similarity $J$ and boundary accuracy $F$. We also report recovery over the first one, five, and ten frames after return as Rec@1, Rec@5, and Rec@10. Target transfer measures whether the predicted mask overlaps a different annotated object more strongly than the intended target. Because only the first read at reappearance is changed, later recovery also reflects the model's return to its intact memory state. Additional substitution controls, robustness tests, and SAM 2 intervention checks are reported in the Appendix.
\vspace{-1em}
\section{World Models Show Limited Exact-Content Dependence}
\label{sec-wm}

Across the world models we study, large memory gains often persist without the exact retrieved content. Frozen latent world models show weak exact-content dependence, while WorldMem shows graded dependence on broader context and exact place.

\subsection{Identity-Free Memory Recovers Most of the Gain}
\label{sec-wm-ladder}

Ego-Exo4D gives the clearest separation between memory benefit and memory specificity. Without memory, revisit consistency error is $0.851$. Correct memory reduces it to $0.456$. Figure~\ref{fig-wm-main}(a) shows that most of this gain does not require the retrieved identity. The global mean of training memories recovers the full benefit. An identity-free barycenter recovers $99\%$, an identity-free sample recovers $97\%$, and a valid memory from another scene recovers $93\%$. In contrast, zero and random controls recover almost none of the gain. A learned constant optimized on training revisits recovers only $40\%$ of the benefit. Thus, even a fixed input trained explicitly to reduce revisit error does not reproduce the full identity-free gain.


The same weak exact-content dependence appears across datasets. Figure~\ref{fig-wm-main}(b) applies the leakage-safe training-memory mean control to Ego-Exo4D, 7-Scenes, and TUM. It recovers $102\%$ of the correct-memory benefit on Ego-Exo4D and $101\%$ on 7-Scenes. The 7-Scenes memory is trained and evaluated on disjoint scenes, so the control contains no information from the evaluation scene. Values slightly above $100\%$ reflect sampling variation around full recovery. TUM shows a mixed regime. The same control recovers $70\%$ of the benefit, with a bootstrap interval of $[59,94]$. This suggests a larger content-specific residual than on Ego-Exo4D or 7-Scenes, but the interval is wide because the leakage-safe split contains only two evaluation takes. We therefore treat TUM as suggestive of mixed content dependence rather than as a precise estimate of the content-specific share.



\subsection{Recovery Falls Away from Observed Training Memories}
\label{sec-wm-dose}

Benefit recovery decreases as an identity-free value moves away from observed training-memory representations. We test this with the dose response defined in Sec.~\ref{sec-method-repair}. Starting from an identity-free value constructed only from training memories, we progressively perturb it away from observed training memories.

On Ego-Exo4D, recovery falls monotonically from $102\%$ at $d_{\mathrm{real}}=0.002$ to $1\%$ at $d_{\mathrm{real}}=0.97$ (Figure~\ref{fig-wm-main}(c)). Evaluation-specific identity is absent throughout the intervention. The changing factor is proximity to memory representations observed during training. This pattern supports representation repair as an explanation for the identity-independent gain in this setting.


\subsection{Direction Matters at Fixed Magnitude}
\label{sec-wm-direction}

The dose response could reflect changes in injection strength rather than representation content. We therefore hold the injected norm fixed and change only its direction.

Recovery still collapses. On Ego-Exo4D, it falls from $101\%$ to $29\%$, then $2\%$, and finally $0\%$. TUM shows the same qualitative pattern, falling from $65\%$ to $0\%$ (Table~\ref{tab-wm-direction}). Injection strength alone is therefore insufficient to explain the identity-independent gain.

\begin{table}[t]
\footnotesize
\centering
\setlength{\tabcolsep}{4pt}
\begin{tabular}{lccccc}
\toprule
Rotation $\alpha$ & $0.00$ & $0.25$ & $0.50$ & $0.75$ & $1.00$ \\
\midrule
Ego-Exo4D & $101.4\%$ & $29.2\%$ & $1.8\%$ & $0.1\%$ & $-0.3\%$ \\
TUM & $65.3\%$ & $14.2\%$ & $-3.0\%$ & $-1.8\%$ & $-0.3\%$ \\
\bottomrule
\end{tabular}
\caption{Fixed-magnitude direction control. The injected norm is fixed while its direction is rotated by $\alpha$. Benefit recovery drops sharply on both datasets.}
\label{tab-wm-direction}
\end{table}
\subsection{Weak Content Dependence Transfers to a Second Host}
\label{sec-wm-host}

The same dissociation appears with an action-conditioned V-JEPA 2 host \cite{vjepa2}. Without memory, revisit consistency error is $0.1312$. Correct memory reduces it to $0.0442$, while wrong memory gives $0.0455$. The wrong-minus-correct RCE gap is $0.0013$, or about $1.5\%$ of the total memory benefit.

We use this experiment only as a second-host check. The full substitution ladder and representation-repair analysis remain restricted to DINO-WM. The V-JEPA 2 result shows that a large memory gain can coexist with weak dependence on the retrieved identity on a second host.

\subsection{WorldMem Shows Graded Content Dependence}
\label{sec-wm-worldmem}

Limited exact-content dependence is not restricted to the memory pathway attached to the frozen hosts. We apply the same audit to WorldMem \cite{xiao2025worldmem}, whose memory is native to a trained generative world model.

\begin{figure}[t]
\centering
\includegraphics[width=\linewidth]{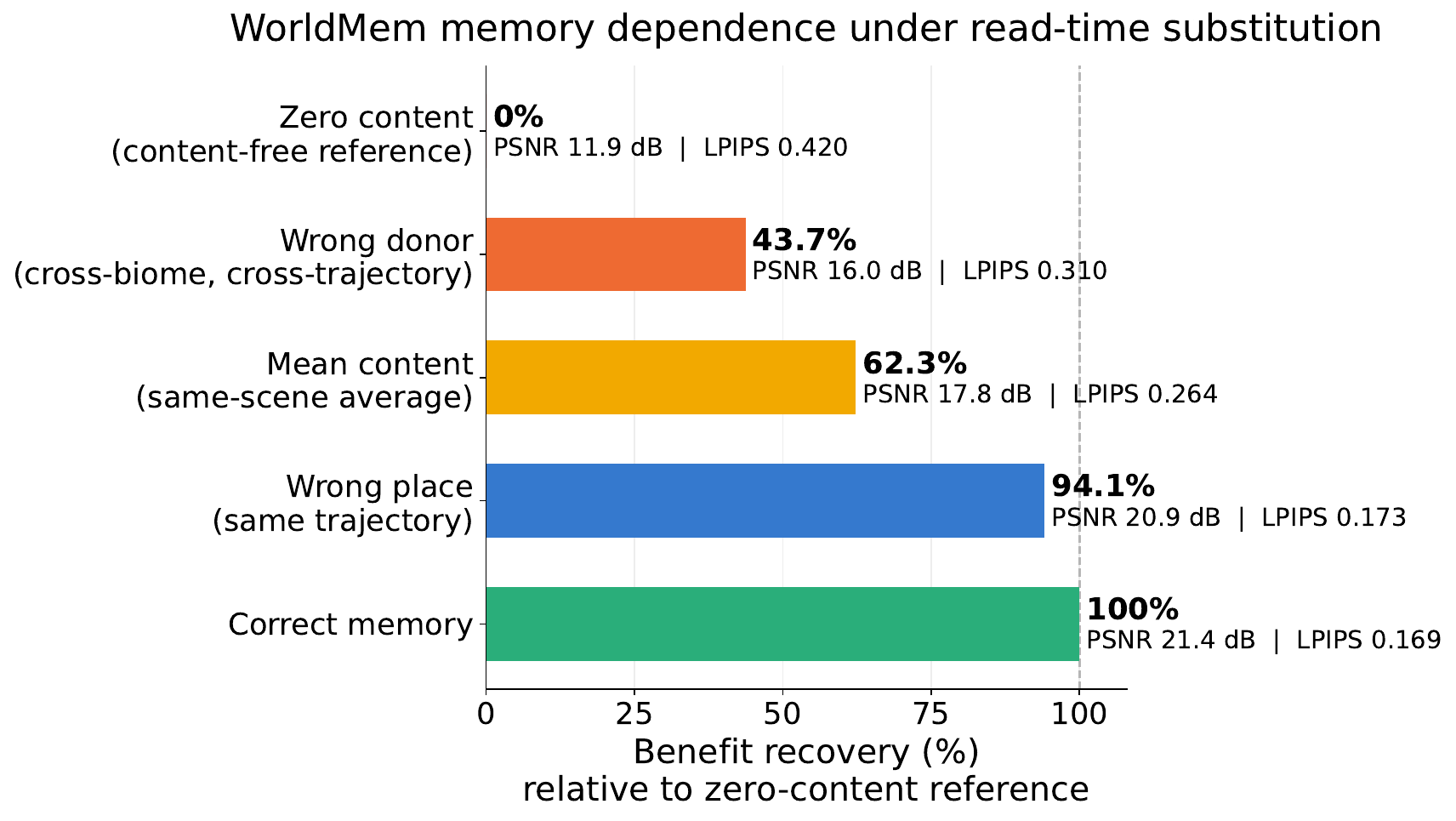}
\caption{Read-time substitution on WorldMem over $1000$ evaluation cases. Bars show benefit recovery relative to the zero-content reference, with PSNR $\uparrow$ and LPIPS $\downarrow$ reported alongside each condition. Recovery increases as more relevant context is preserved. A donor from a disjoint trajectory and biome retains $43.7\%$ of the reference-relative benefit, while a wrong place from the same trajectory retains $94.1\%$.}
\label{fig-worldmem}
\end{figure}

Figure~\ref{fig-worldmem} reports results over $1000$ evaluation cases. All recovery values are measured relative to the zero-content reference $S_0$. Correct memory achieves a PSNR of $21.4$. A wrong memory from another place in the same trajectory gives $20.9$ PSNR and recovers $94.1\%$ of the reference-relative benefit. Exact place identity therefore adds little beyond a memory that preserves the broader trajectory context.

Removing more context produces a larger drop. A valid donor from a disjoint trajectory and different biome gives $16.0$ PSNR and recovers $43.7\%$. Mean content from the same scene gives $17.8$ PSNR and recovers $62.3\%$. LPIPS follows the same ordering.

Preserving broader trajectory context retains much more of the benefit, while exact place identity adds only a small residual beyond the same-trajectory control. This experiment does not establish the representation-repair mechanism measured on DINO-WM. It shows that limited exact-content dependence also occurs in a native generative memory system, where broader contextual information remains important.

\section{SAM 2 Shows Strong Content Dependence}
\label{sec-sam2}

SAM 2 shows a different pattern from the world models. Its segmentation output depends strongly on the content of the spatial memory, both when the target is visible and when it reappears after disappearance.

\boldheader{Spatial Memory Is Strongly Content Dependent on DAVIS. } On DAVIS 2017, replacing the spatial memory removes most of the segmentation performance. Correct memory gives $\mathcal{J}\&\mathcal{F}=0.926$, while wrong spatial memory gives $0.182$, an identity-free prototype gives $0.126$, and zero memory gives $0.173$. Replacing both memory channels gives a similar score of $0.183$. In contrast, replacing only the object pointer leaves performance unchanged at $0.927$ (Table~\ref{tab-sam2-davis}). The spatial-memory collapse occurs clearly in eight of ten sequences. In the remaining two, current-frame evidence is sufficient to preserve the target despite the substitution. The object-pointer result is specific to this setting. A valid wrong pointer has almost no effect on these generally visible target frames.


\begin{table}[t]
\footnotesize
\centering
\begin{tabular}{lc}
\toprule
Read-time substitution & $\mathcal{J}\&\mathcal{F}$ \\
\midrule
Correct memory & $0.926$ \\
Wrong spatial memory & $0.182$ \\
Wrong object pointer & $0.927$ \\
Wrong spatial memory and pointer & $0.183$ \\
Identity-free prototype & $0.126$ \\
Zero memory & $0.173$ \\
\bottomrule
\end{tabular}
\caption{Read-time memory substitution on DAVIS 2017 over ten sequences. Replacing the spatial memory with a valid memory from another object removes most of the segmentation performance, while replacing the object pointer has little effect at these target frames.}
\label{tab-sam2-davis}
\end{table}

\begin{figure}[t]
\centering
\includegraphics[width=\linewidth]{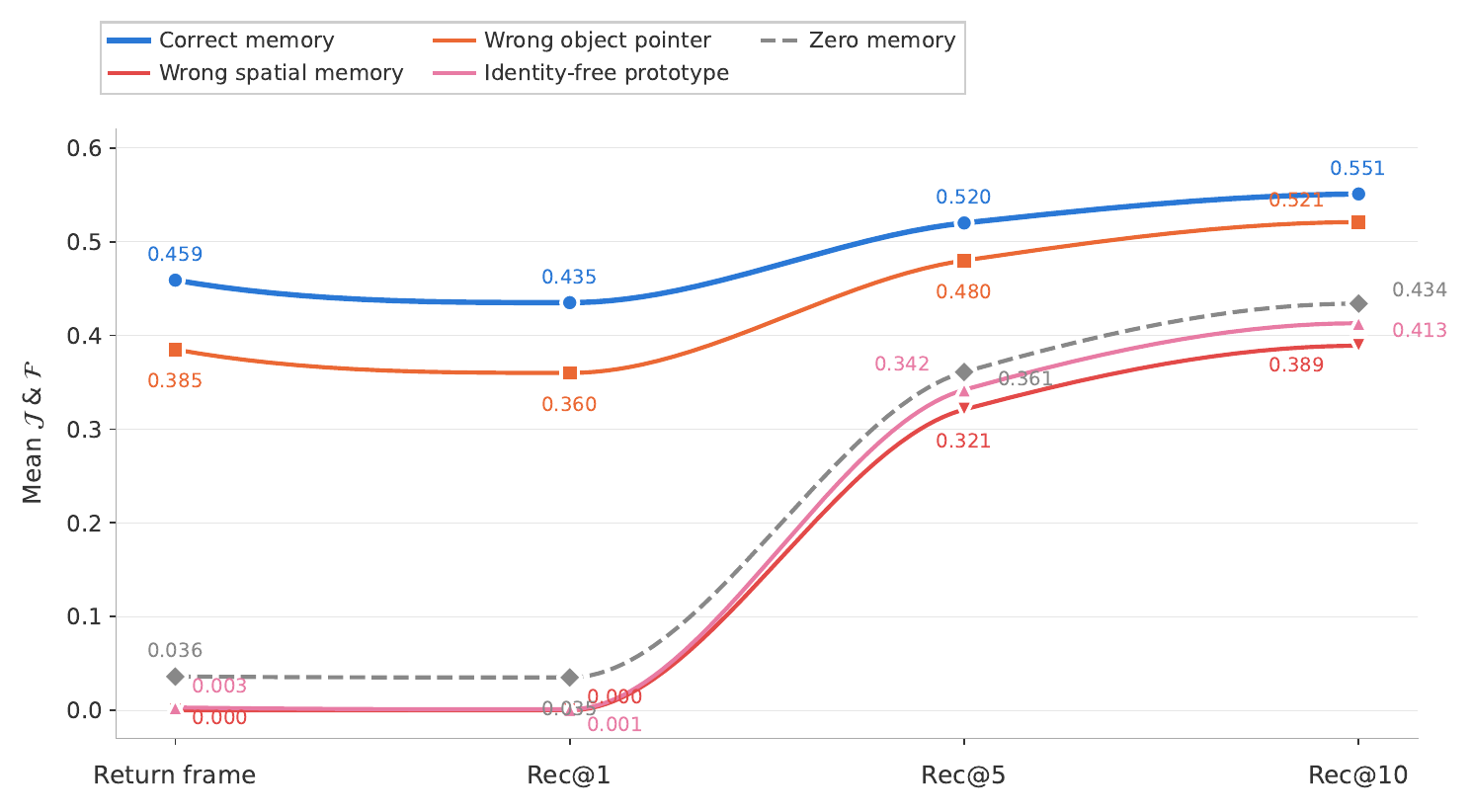}
\caption{Read-time memory substitution at MOSEv2 reappearance over $100$ events. $J$ and $F$ are measured at return, and Rec@$k$ reports mean region agreement over the first $k$ frames after return. Spatial-memory substitution causes the largest collapse, while the object-pointer substitution has a smaller effect.}
\label{fig-sam2-mose}
\end{figure}

\boldheader{Correct Spatial Memory Is Critical at Reappearance.} Spatial-memory dependence becomes stronger when the target must be recovered after disappearance. Across $100$ MOSEv2 reappearance events, correct memory gives a mean $J$ and $F$ score of $0.459$ at the return frame. Wrong spatial memory reduces the score to $0.000$, while an identity-free prototype gives $0.003$ and zero memory gives $0.036$ (Figure~\ref{fig-sam2-mose}). The effect is strongest at reappearance. Performance partially recovers over later frames because only the first memory read at the return frame is changed, after which the model resumes using its intact memory state. The object pointer also contributes in this harder setting. Replacing it reduces the return-frame score from $0.459$ to $0.385$, and among the $49$ events where SAM 2 re-detects the object with correct memory, performance falls from $0.910$ to $0.757$. The pointer is therefore partially load bearing at reappearance despite having little effect on the DAVIS frames. Target transfer remains at or near zero across all conditions, indicating that spatial-memory corruption usually causes the target to be lost rather than transferred to the donor object. Together, these results show strong spatial-memory content dependence at reappearance, with a smaller but measurable contribution from the object pointer.

\section{Discussion and Conclusion}
\label{sec-discussion}

\textbf{What memory benefit means.} Our results show that memory benefit and memory specificity capture different properties. Memory can help through generic representation support, broader contextual information, or exact episodic content. Ego-Exo4D and 7-Scenes show limited exact-content dependence, TUM suggests a mixed regime, WorldMem shows graded dependence on context, and SAM 2 depends strongly on correct spatial memory. These regimes imply different design choices. If identity-free memory preserves most of the gain, better retrieval cannot address that component and effort may be better directed toward representation stability or predictor quality. If broader context matters, retrieval should preserve the relevant scene or trajectory context. When exact content is load bearing, retrieval correctness becomes critical. The DINO-WM interventions further support representation repair as one mechanism for low content dependence because recovery falls with distance from observed training-memory representations and with direction changes at fixed magnitude. WorldMem does not establish the same mechanism, but shows that a native memory system can depend strongly on broader context while depending little on the exact place. We therefore recommend evaluating memory benefit together with a substitution ladder rather than relying on a single memory-on versus memory-off score.

\textbf{Implications and future directions.} Read-time substitution can serve as a diagnostic before investing in more complex retrieval. Comparing correct memory, context-matched and context-mismatched wrong memories, identity-free controls, and a content-free reference reveals whether a system needs exact retrieval, broader context, or generic memory support. The desired specificity is likely task dependent. Long-range generation may require scene consistency without exact frame retrieval, while object recovery after occlusion may require highly specific stored content. This raises concrete research questions about what task ambiguity, training objectives, memory bottlenecks, temporal distance, and current evidence cause specificity to emerge. Future work can design objectives that explicitly reward the desired level of memory use and apply interventions across channels or layers to localize where stored information becomes causally important. The same principle can extend to long-video language models, generative world models, embodied agents, recurrent state, KV caches, long-context attention, and learned memory tokens.

\boldheader{Limitations. }Our conclusions are bounded by the models and interventions studied here. The full substitution ladder, dose response, and fixed-magnitude analysis are performed on DINO-WM, while V-JEPA 2 provides only a correct-versus-wrong second-host check. WorldMem extends the audit to a native generative memory system, but we do not run the same representation-repair analysis on that model. Its controls also change several forms of context at once, so we interpret them as evidence of graded trajectory and scene dependence rather than isolate a single semantic factor. Because WorldMem does not support removing its memory interface directly, recovery is measured relative to zero content rather than a true memory-off condition. The leakage-safe TUM split contains only two evaluation takes, so its result is suggestive of mixed content dependence rather than a precise estimate of the content-specific share. Our $d_{\mathrm{real}}$ measure is an empirical distance to observed training-memory representations, not a formal manifold distance, and revisit consistency error measures latent consistency rather than perceptual generation quality or downstream control. SAM 2 provides one native-memory example and should not be taken as representative of all end-to-end memory systems. Our intervention changes one memory read, so it measures direct dependence at the target step. On MOSEv2, later recovery partly reflects the return to intact memory, and the training split is used because dense masks are required to identify and score reappearance events. The audit also requires access to the memory-read interface and is less direct for black-box or implicit memory systems. Finally, because the model families use different tasks and metrics, we compare patterns of content dependence rather than assign a common numerical specificity ranking across systems.

\boldheader{Conclusion. }We asked whether the benefit of memory in video models depends on what is retrieved. Read-time memory substitution answers this by changing the consumed memory content while preserving the surrounding computation. Frozen world models show limited exact-content dependence, and the Ego-Exo4D interventions support representation repair as the best-supported explanation for the identity-independent gain in that setting. WorldMem extends the finding to a natively trained generative memory system with graded content dependence, where broader trajectory context matters substantially but the exact past place adds little beyond it. SAM 2 provides the contrasting case, with strong dependence on the correct spatial memory including when an object must be recovered after disappearance. Memory benefit and memory specificity are therefore different properties, and memory gains can depend on stored information at different levels. Measuring this dependence reveals whether progress should target representation stability, contextual retrieval, or precise recovery of past information. A model should not be said to recall the past merely because memory improves its output. The improvement should depend on what was retrieved.



\clearpage
\newpage
{
    \small
    \bibliographystyle{ieeenat_fullname}
    \bibliography{main}
}
\clearpage
\appendix

\section*{Appendix}
\section*{Impact Statement}

Memory is increasingly used to extend video models over long sequences, but performance gains alone do not reveal what information the model actually uses. Our audit provides a direct way to distinguish gains that depend on retrieved content from gains that arise through broader contextual or representation-level effects. This distinction can change how memory systems are evaluated and designed. If performance does not depend on the retrieved identity, additional retrieval capacity, larger memory stores, or more complex indexing may provide limited value for that part of the gain. If exact content is load bearing, retrieval quality becomes critical for reliability. More broadly, measuring memory specificity can help separate genuine use of stored evidence from improvements that only appear to come from recall. This provides a practical diagnostic for video models and a general direction for studying memory use in world models, embodied agents, and other long-context systems.

\section{Supplementary Index}
\label{app-index}

The supplementary material provides implementation details, additional mechanism tests, robustness analyses, and the full numerical results behind the main-paper findings. Table~\ref{tab-app-index} summarizes the role of each section.

\begin{table}[h]
\footnotesize
\centering
\begin{tabular}{p{0.19\columnwidth}p{0.70\columnwidth}}
\toprule
Location & Purpose \\
\midrule
Sec.~\ref{app-details} & Defines the intervention points and construction of the memory controls. \\
Sec.~\ref{app-expanded-ladder} & Reports the full Ego-Exo4D substitution ladder. \\
Sec.~\ref{app-cross-dataset} & Tests identity-free recovery across datasets. \\
Sec.~\ref{app-dose-diagnostics} & Tests recovery as values move away from observed training memories. \\
Sec.~\ref{app-fixed-magnitude} & Tests whether injection magnitude alone explains recovery. \\
Sec.~\ref{app-gate} & Tests whether differential gate activation explains identity-independent recovery. \\
Sec.~\ref{app-multiread} & Tests whether low specificity persists across repeated memory reads. \\
Sec.~\ref{app-memory-config} & Tests robustness to retrieval width and memory budget. \\
Sec.~\ref{app-scene-appearance} & Tests whether scene appearance explains the weak Ego-Exo4D identity gap. \\
Sec.~\ref{app-vjepa} & Reports additional identity-free controls for the V-JEPA 2 second-host check. \\
Sec.~\ref{app-worldmem} & Reports the final $1000$-case WorldMem content ladder. \\
Sec.~\ref{app-sam2-davis} & Reports spatial-memory specificity on DAVIS. \\
Sec.~\ref{app-sam2-verification} & Verifies that the SAM 2 interventions alter the intended pathways. \\
Sec.~\ref{app-sam2-mose} & Reports the full MOSEv2 reappearance results. \\
Sec.~\ref{app-sam2-bootstrap} & Adds uncertainty estimates for the MOSEv2 result. \\
Sec.~\ref{app-sam2-redetected} & Isolates the object-pointer contribution after successful re-detection. \\
Sec.~\ref{app-sam2-duration} & Tests spatial-memory dependence across disappearance durations. \\
\bottomrule
\end{tabular}
\caption{Index of the supplementary analyses and the question addressed by each one.}
\label{tab-app-index}
\end{table}

\section{Intervention and Control Details}
\label{app-details}

The main paper defines read-time memory substitution at the consumed memory value. We provide the exact intervention points and control constructions here.

\subsection{Intervention Locations}
\label{app-intervention-locations}

The intervention changes the memory representation consumed at the target read. The query, preceding memory trajectory, and model weights remain unchanged. Table~\ref{tab-app-intervention-locations} summarizes the intervention point for each system.

\begin{table*}[t]
\scriptsize
\centering
\setlength{\tabcolsep}{2.5pt}
\renewcommand{\arraystretch}{1.05}
\begin{tabular}{
p{0.12\textwidth}
p{0.19\textwidth}
p{0.22\textwidth}
p{0.16\textwidth}
p{0.08\textwidth}
p{0.10\textwidth}
}
\toprule
System & Replaced component & Intervention point & Position in pathway & Query fixed & Earlier trajectory fixed \\
\midrule
DINO-WM
& Memory read-output vector
& Context value supplied to the memory injector
& Before projection and gate
& Yes
& Yes \\

V-JEPA 2
& Memory read-output vector
& Same memory-adapter interface as DINO-WM
& Before projection and gate
& Yes
& Yes \\

WorldMem
& Retrieved frame latents
& Gathered memory latents
& Before memory attention
& Yes
& Yes \\

SAM 2 spatial
& \texttt{maskmem\_features}
& \shortstack[l]{\texttt{\_prepare\_memory\_}\\\texttt{conditioned\_features}}
& Before memory attention
& Yes
& Yes \\

SAM 2 pointer
& \texttt{obj\_ptr}
& Same read-time hook
& Before memory attention
& Yes
& Yes \\
\bottomrule
\end{tabular}
\caption{Read-time intervention locations across model families. Only the consumed memory representation is replaced. The query and preceding memory trajectory remain fixed.}
\label{tab-app-intervention-locations}
\end{table*}


These interventions preserve the native memory pathway while changing the information supplied through it. The audit therefore measures dependence on consumed memory content rather than the effect of removing the memory architecture.

\subsection{Construction of DINO-WM Memory Controls}
\label{app-control-construction}

The identity-free controls contain no evaluation-specific identity. Training-derived controls use training data only. Table~\ref{tab-app-control-construction} gives the construction and information available to each control.


\begin{table*}[t]
\scriptsize
\centering
\setlength{\tabcolsep}{2.5pt}
\renewcommand{\arraystretch}{1.05}
\begin{tabular}{
p{0.13\textwidth}
p{0.30\textwidth}
p{0.13\textwidth}
p{0.09\textwidth}
p{0.11\textwidth}
p{0.09\textwidth}
}
\toprule
Control
& Construction
& Source
& Query dependent
& Uses current eval trajectory
& Uses eval identity \\
\midrule
Correct memory
& Native pose-keyed read for the current query
& Evaluation trajectory
& Yes
& Yes
& Yes \\

Wrong memory
& Valid memory read from another scene
& Donor scene
& No
& No
& No \\

Training-memory mean
& Mean of pooled memory read outputs
& Training reads only
& No
& No
& No \\

Identity-free barycenter
& Mean of sampled training read outputs, norm matched to the median read norm
& Training reads only
& No
& No
& No \\

Identity-free sample
& PCA-Gaussian sample in memory-read space, norm matched to valid reads
& Training reads only
& No
& No
& No \\

Learned constant
& One free $384$-D vector optimized on training revisit pairs through the frozen pathway
& Training revisit pairs
& No
& No
& No \\

Random
& Gaussian direction with a matched reference norm
& None
& No
& No
& No \\

Zero
& All zeros
& None
& No
& No
& No \\
\bottomrule
\end{tabular}
\caption{Construction of the DINO-WM substitution controls. Training-derived identity-free controls use training data only and contain no evaluation-specific identity.}
\label{tab-app-control-construction}
\end{table*}

The controls remove different information. Wrong memory preserves a valid memory representation but changes its identity. Identity-free controls remove evaluation identity while preserving selected properties of training-memory representations. Random and zero controls do not preserve the same memory structure.

\section{Additional DINO-WM Evidence}
\label{app-dino}

Each experiment below tests a different explanation for the observed identity-independent gain.

\subsection{Expanded Ego-Exo4D Substitution Ladder}
\label{app-expanded-ladder}

Most of the Ego-Exo4D memory benefit survives without the correct retrieved identity. We report the full ladder over $933$ revisits from $39$ takes in Table~\ref{tab-app-expanded-ladder}. Confidence intervals use a cluster bootstrap over takes.

\begin{table}[t]
\footnotesize
\centering
\begin{tabular}{lc}
\toprule
Injected memory value & Benefit recovery \\
\midrule
Zero & $0\%$ \\
Random & $1\%$ \\
Learned constant & $40\%$ \\
Wrong memory, other scene & $93\%$ \\
Identity-free sample & $97\%\ [94,101]$ \\
Identity-free barycenter & $99\%\ [96,101]$ \\
Training-memory mean & $100\%$ \\
Correct memory & $100\%$ \\
\bottomrule
\end{tabular}
\caption{Expanded read-time substitution ladder on Ego-Exo4D with DINO-WM over $933$ revisits from $39$ takes.}
\label{tab-app-expanded-ladder}
\end{table}

The training-memory mean recovers the full benefit, while the identity-free controls recover nearly all of it. A valid memory from another scene also recovers $93\%$. Random and zero controls recover almost none. The learned constant recovers only $40\%$, showing that full recovery is not produced by every fixed input to the memory pathway.

\subsection{Leakage-Safe Recovery Across Datasets}
\label{app-cross-dataset}

The weak exact-content dependence is not limited to Ego-Exo4D. We apply the leakage-safe training-memory mean control to Ego-Exo4D, 7-Scenes, and TUM. Table~\ref{tab-app-cross-dataset} reports the final recovery values.

\begin{table}[t]
\footnotesize
\centering
\begin{tabular}{lcc}
\toprule
Dataset & Recovery & Residual relative to control \\
\midrule
Ego-Exo4D & $102\%$ & $\sim0\%$ \\
7-Scenes & $101\%$ & $\sim0\%$ \\
TUM & $70\%\ [59,94]$ & $\sim30\%$ \\
\bottomrule
\end{tabular}
\caption{Leakage-safe training-memory mean recovery across datasets. The TUM split contains two evaluation takes.}
\label{tab-app-cross-dataset}
\end{table}

Ego-Exo4D and 7-Scenes show little dependence on evaluation-specific identity under this control. Values slightly above $100\%$ are un-clipped finite-sample estimates around full recovery. TUM shows more evidence of mixed dependence, but the leakage-safe split contains only two evaluation takes. We therefore treat TUM as evidence that mixed dependence is possible rather than as a precise estimate of the content-specific share.

\subsection{Dose Response and Internal Diagnostics}
\label{app-dose-diagnostics}

Recovery falls as the training-memory mean is progressively perturbed away from memory representations observed during training. As defined in Eq.~(3), $d_{\mathrm{real}}$ is the cosine distance to the nearest observed training-memory representation. The reference bank contains training memories only. We use this quantity as an empirical support measure and do not interpret it as distance to a formal representation manifold.

\begin{figure*}[t]
\centering
\includegraphics[width=\textwidth]{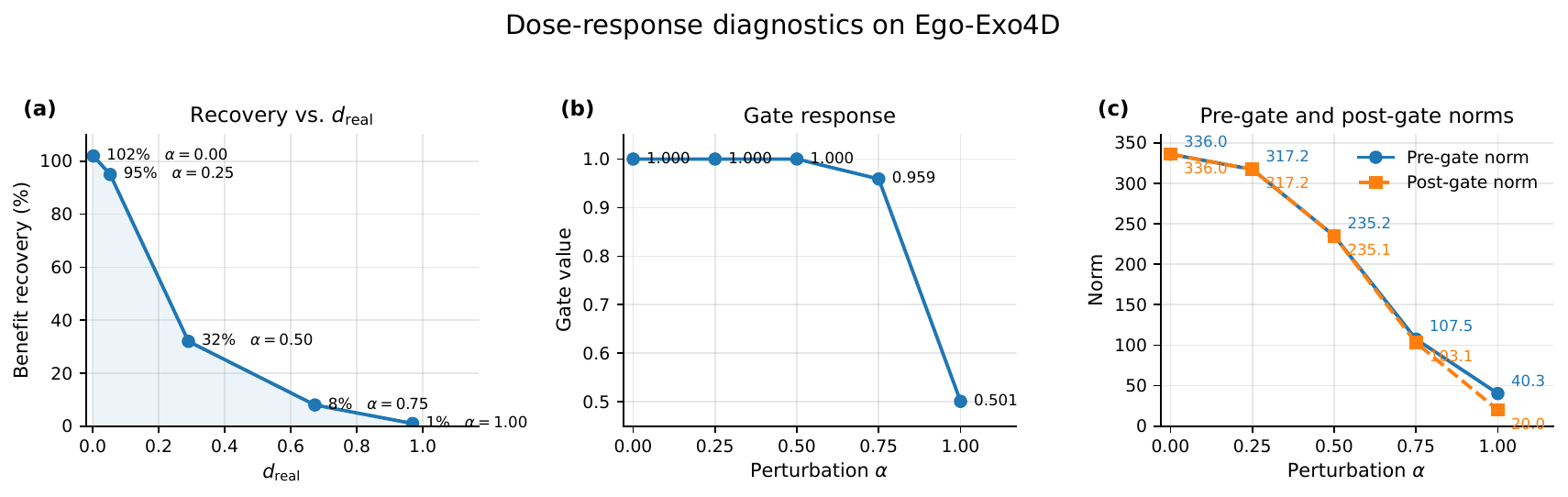}
\caption{Dose-response diagnostics on Ego-Exo4D. (a) Benefit recovery decreases as the perturbed training-memory mean moves farther from observed training-memory representations. (b) The gate remains fully open through $\alpha=0.50$, even as recovery falls to $32\%$. (c) Pre-gate and post-gate norms decrease at larger perturbations. The early recovery drop therefore cannot be explained by differential gate activation, while Table~\ref{tab-app-fixed-magnitude} separately controls for injection magnitude.}
\label{fig-app-dose}
\end{figure*}

Figure~\ref{fig-app-dose} adds gate and norm diagnostics to the dose response reported in the main paper.


Recovery has already fallen from $102\%$ to $32\%$ at $\alpha=0.50$, while the gate remains fully open at $1.000$. Differential gate activation therefore does not explain the early decline. At larger perturbations, both gate activation and injected magnitude change. We therefore test magnitude separately.

\subsection{Direction at Fixed Injection Magnitude}
\label{app-fixed-magnitude}

We hold injection magnitude fixed and change only its direction. The norm is fixed to the median norm of real memory-induced deltas. This is $335.0$ on Ego-Exo4D and $346.6$ on TUM.

Recovery still collapses on both datasets, as shown in Table~\ref{tab-app-fixed-magnitude}.

\begin{table}[t]
\footnotesize
\centering
\setlength{\tabcolsep}{4pt}
\begin{tabular}{ccc}
\toprule
Rotation $\alpha$ & Ego-Exo4D & TUM \\
\midrule
$0.00$ & $101.4\%$ & $65.3\%$ \\
$0.25$ & $29.2\%$ & $14.2\%$ \\
$0.50$ & $1.8\%$ & $-3.0\%$ \\
$0.75$ & $0.1\%$ & $-1.8\%$ \\
$1.00$ & $-0.3\%$ & $-0.3\%$ \\
\bottomrule
\end{tabular}
\caption{Benefit recovery when injection magnitude is fixed and only direction is changed. The injected norm is $335.0$ on Ego-Exo4D and $346.6$ on TUM.}
\label{tab-app-fixed-magnitude}
\end{table}

Recovery drops despite constant injection magnitude. Injection strength alone is therefore insufficient to explain the identity-independent gain. Negative values are retained because benefit recovery is not clipped.

\subsection{Gate and Injection Diagnostics}
\label{app-gate}

We next test whether valid identity-free memories recover the benefit because the learned gate treats them differently. This diagnostic uses $100$ Ego-Exo4D revisits and records the scalar gate and injected delta norm.

Valid high-recovery substitutions all keep the gate fully open and produce similar injected magnitudes, as shown in Table~\ref{tab-app-gate}.

\begin{table*}[t]
\centering

\begin{minipage}[t]{0.49\textwidth}
\centering
\scriptsize
\setlength{\tabcolsep}{3pt}

\resizebox{\linewidth}{!}{%
\begin{tabular}{lccccc}
\toprule
Condition
& RCE $\downarrow$
& Recovery
& Gate
& Pre-gate $\lVert\Delta\rVert$
& Post-gate $\lVert\Delta\rVert$ \\
\midrule
Correct
& $0.444$
& $100\%$
& $1.000$
& $328.5$
& $328.5$ \\

Wrong scene
& $0.447$
& $99\%$
& $1.000$
& $329.5$
& $329.5$ \\

Training-memory mean
& $0.433$
& $103\%$
& $1.000$
& $328.7$
& $328.7$ \\

Identity-free barycenter
& $0.432$
& $103\%$
& $1.000$
& $329.4$
& $329.4$ \\

Identity-free sample
& $0.432$
& $103\%$
& $1.000$
& $333.5$
& $333.5$ \\

Learned constant
& $0.699$
& $36\%$
& $1.000$
& $79.7$
& $79.6$ \\

Random
& $0.843$
& $0\%$
& $0.586$
& $23.6$
& $13.8$ \\

Zero
& $0.845$
& $0\%$
& $0.499$
& $1.1$
& $0.5$ \\
\bottomrule
\end{tabular}%
}

\captionof{table}{
Gate and injection diagnostics on $100$ Ego-Exo4D revisits.
This instrumented diagnostic is separate from the $933$-revisit main evaluation.
}
\label{tab-app-gate}

\end{minipage}
\hfill
\begin{minipage}[t]{0.49\textwidth}
\centering
\scriptsize
\setlength{\tabcolsep}{3pt}

\resizebox{\linewidth}{!}{%
\begin{tabular}{lccccc}
\toprule
Read-time substitution
& $\mathcal{J}$\&$\mathcal{F}$
& Rec@1
& Rec@5
& Rec@10
& Target transfer \\
\midrule
Correct memory
& $0.459$
& $0.435$
& $0.520$
& $0.551$
& $0.0\%$ \\

Wrong spatial memory
& $0.000$
& $0.000$
& $0.321$
& $0.389$
& $0.0\%$ \\

Wrong object pointer
& $0.385$
& $0.360$
& $0.480$
& $0.521$
& $0.0\%$ \\

Wrong spatial memory and pointer
& $0.000$
& $0.000$
& $0.321$
& $0.389$
& $0.0\%$ \\

Identity-free prototype
& $0.003$
& $0.001$
& $0.342$
& $0.413$
& $0.0\%$ \\

Zero memory
& $0.036$
& $0.035$
& $0.361$
& $0.434$
& $1.0\%$ \\
\bottomrule
\end{tabular}%
}

\captionof{table}{Read-time substitution at object reappearance on MOSEv2 over $100$ events. $\mathcal{J}\&\mathcal{F}$ is measured at the return frame. Rec@$k$ reports mean region agreement over the first $k$ frames after return. Target transfer is the fraction of events where the predicted mask overlaps another annotated object more strongly than the intended target.}
\label{tab-app-sam2-mose}

\end{minipage}

\end{table*}

Correct, wrong-scene, training-mean, barycenter, and identity-free sample memories all keep the gate fully open and produce injected norms of approximately $328$ to $334$. Their high recovery is therefore not explained by differential gate activation. Random and zero inputs produce much smaller signals. The fixed-magnitude direction control in Table~\ref{tab-app-fixed-magnitude} separately shows that injection magnitude alone is insufficient.

\subsection{Specificity Under Repeated Memory Reads}
\label{app-multiread}

The main audit changes one memory read and measures direct dependence at that step. We additionally test whether low specificity persists when several consecutive reads are replaced. We substitute $k\in\{1,3,5,10\}$ consecutive reads on an Ego-Exo4D diagnostic set of $100$ revisits. Recovery is measured at the final substituted read. The identity-free value is the training-memory mean, computed as the mean of memory read outputs collected over the training trajectories and injected in the read-output space.

Low specificity persists through repeated intervention, as shown in Figure~\ref{fig-app-multiread}. The training-memory mean retains approximately full recovery through ten substituted reads. Wrong-scene recovery decreases from $102\%$ to $89\%$, showing that repeatedly injecting conflicting scene content introduces a modest cumulative penalty. Zero memory remains at $0\%$.

This experiment complements the single-read audit rather than replacing it. The main audit measures direct dependence at one target read. The repeated-read experiment also includes effects that accumulate after earlier substitutions. This experiment is a separate diagnostic run from the gate analysis, so its single-read values are reported independently.

\subsection{Robustness to Retrieval Width and Memory Budget}
\label{app-memory-config}

Weak correct-versus-wrong dependence could result from one retrieval width or memory budget. We vary both quantities and measure the direct RCE gap between correct and wrong memory. This analysis is separate from the leakage-safe identity-free recovery result.

\begin{figure}[t]
\centering
\includegraphics[width=\linewidth]{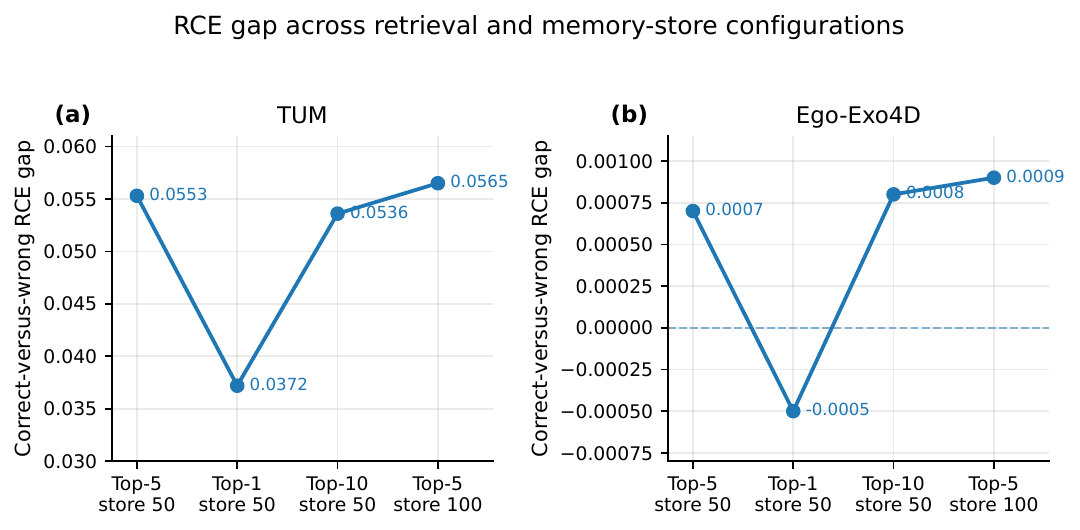}
\caption{Correct-versus-wrong revisit consistency error (RCE) gap under different retrieval widths and memory-store sizes. Ego-Exo4D remains near zero across the tested configurations, while TUM retains a larger gap. The qualitative dataset difference is therefore not tied to one retrieval width or memory budget.}
\label{fig-app-memory-config}
\end{figure}

The qualitative dataset difference persists across the tested configurations, as shown in Figure~\ref{fig-app-memory-config}.


Ego-Exo4D retains a near-zero correct-versus-wrong gap across the tested configurations. TUM retains a larger gap. The qualitative difference is therefore not tied to one retrieval width or memory budget.

\subsection{Robustness to Scene Appearance}
\label{app-scene-appearance}

One possible explanation for weak specificity on Ego-Exo4D is that visually similar scenes make different memories interchangeable. We split takes at the activity-scenario level. The distinctive group contains cooking, sushi, salad, omelet preparation, music, and dance scenarios. The generic group contains basketball, soccer, biking, CPR, and COVID testing scenarios. This gives $111$ distinctive and $108$ generic takes. We train the same memory pathway with three seeds for each group.

\begin{figure*}[t]
\centering

\begin{minipage}[t]{0.49\textwidth}
\vspace{0pt}
\centering

\includegraphics[
    width=\linewidth,
    height=0.60\textwidth,
    keepaspectratio
]{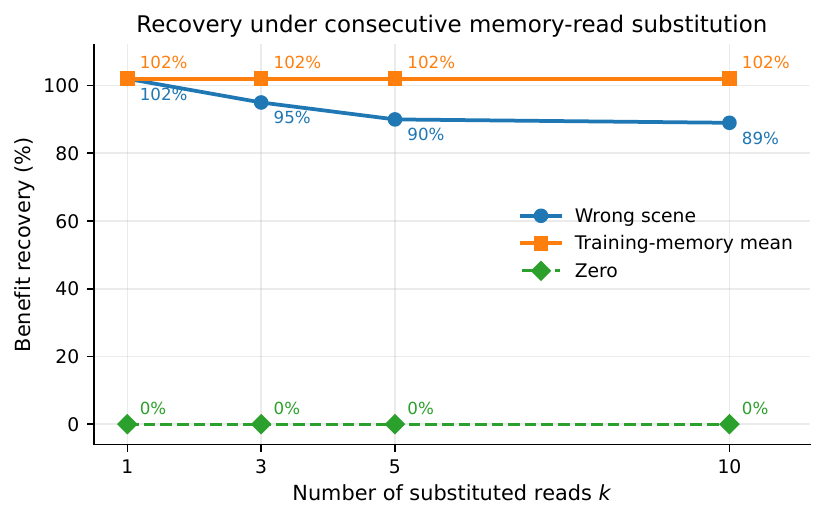}

\captionof{figure}{
Benefit recovery after replacing $k$ consecutive memory reads on an Ego-Exo4D diagnostic set of $100$ revisits. Recovery is measured at the final substituted read. The training-memory mean retains approximately full recovery across repeated reads, while wrong-scene recovery decreases modestly as the intervention is sustained. Zero memory remains at $0\%$ throughout.
}
\label{fig-app-multiread}

\end{minipage}
\hfill
\begin{minipage}[t]{0.49\textwidth}
\vspace{0pt}
\centering

\includegraphics[
    width=\linewidth,
    height=0.60\textwidth,
    keepaspectratio
]{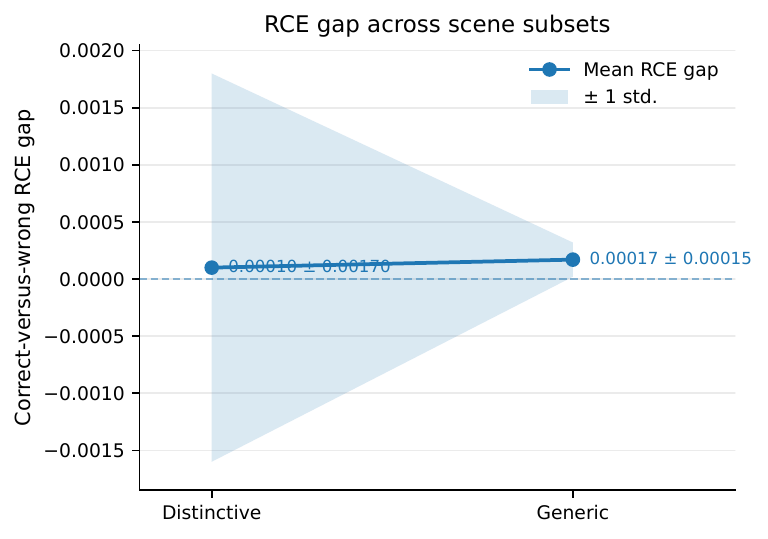}

\captionof{figure}{Correct-versus-wrong revisit consistency error (RCE) gap on visually distinctive and generic Ego-Exo4D subsets. Values summarize the mean across three seeds, with variability shown as one standard deviation. The two groups show similarly small gaps under this partition.}
\label{fig-app-scene-appearance}

\end{minipage}

\end{figure*}

The two groups show similarly small correct-versus-wrong gaps, as reported in Figure~\ref{fig-app-scene-appearance}. Under this scenario-level partition, scene appearance does not explain the weak correct-versus-wrong dependence.





\begin{figure*}[t!]
\centering
\includegraphics[width=\textwidth]{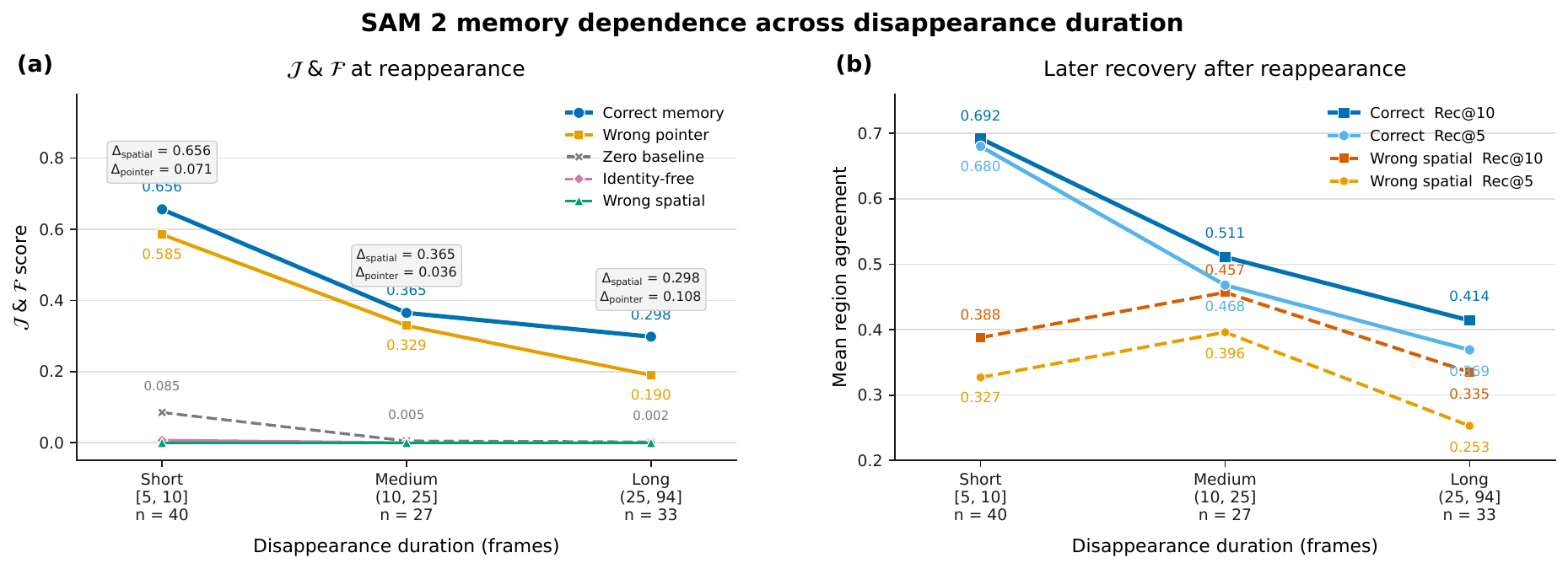}
\caption{
SAM 2 memory dependence across disappearance duration.
(a) $\mathcal{J}$\&$\mathcal{F}$ at reappearance across disappearance-duration groups. Spatial drop is correct minus wrong-spatial performance, and pointer drop is correct minus wrong-pointer performance.
(b) Later recovery with correct and wrong spatial memory, measured by mean region agreement at Rec@5 and Rec@10.
}
\label{fig-sam2-duration}
\end{figure*}



\section{V-JEPA 2 Second-Host Check}
\label{app-vjepa}

The main paper uses V-JEPA 2\cite{vjepa2} as a second-host check based on the correct-versus-wrong memory contrast. We add a supplementary ablation that tests two training-derived identity-free controls, together with random and zero inputs. This is a separate evaluation over $1000$ pairs, so its absolute RCE values differ slightly from the coarse check reported in the main paper.

The additional controls show the same weak dependence on memory identity, as reported in Table~\ref{tab-app-vjepa}.

Correct and wrong memory give the same RCE in this evaluation. The training-memory mean and identity-free barycenter also recover approximately $98\%$ of the correct-memory benefit. Random memory recovers only $26.1\%$, while zero performs slightly worse than the no-memory reference. These controls provide additional evidence that weak dependence on retrieved identity is not specific to DINO-WM.

This ablation does not reproduce the full DINO-WM substitution ladder or its mechanism tests. The learned-constant and identity-free-sample controls, dose response, and fixed-magnitude analysis remain restricted to DINO-WM.

\begin{table*}[t]
\centering

\begin{minipage}[t]{0.32\textwidth}
\vspace{0pt}
\centering
\scriptsize
\setlength{\tabcolsep}{2pt}

\resizebox{\linewidth}{!}{%
\begin{tabular}{lcc}
\toprule
Memory condition & RCE $\downarrow$ & Recovery \\
\midrule
No memory & $0.1309$ & $0\%$ \\
Zero & $0.1356$ & $-5.4\%$ \\
Random & $0.1085$ & $26.1\%$ \\
Training-memory mean & $0.0467$ & $98.1\%$ \\
Identity-free barycenter & $0.0466$ & $98.3\%$ \\
Wrong memory, other scene & $0.0451$ & $100\%$ \\
Correct memory & $0.0451$ & $100\%$ \\
\bottomrule
\end{tabular}%
}

\captionof{table}{
Additional memory-control ablation on the action-conditioned V-JEPA 2 host over $1000$ evaluation pairs. Recovery is measured relative to the no-memory reference.
}
\label{tab-app-vjepa}

\end{minipage}
\hfill
\begin{minipage}[t]{0.32\textwidth}
\vspace{0pt}
\centering
\scriptsize
\setlength{\tabcolsep}{2pt}

\resizebox{\linewidth}{!}{%
\begin{tabular}{p{0.48\linewidth}ccc}
\toprule
Memory condition
& PSNR $\uparrow$
& LPIPS $\downarrow$
& Recovery \\
\midrule
Zero content
& $11.9$
& $0.420$
& $0\%$ \\

Wrong donor, disjoint trajectory and different biome
& $16.0$
& $0.310$
& $43.7\%$ \\

Mean content, same scene
& $17.8$
& $0.264$
& $62.3\%$ \\

Wrong place, same trajectory
& $20.9$
& $0.173$
& $94.1\%$ \\

Correct memory
& $21.4$
& $0.169$
& $100\%$ \\
\bottomrule
\end{tabular}%
}

\captionof{table}{
Read-time substitution on WorldMem over $1000$ evaluation cases. Recovery is computed from PSNR relative to the zero-content reference.
}
\label{tab-app-worldmem}

\end{minipage}
\hfill
\begin{minipage}[t]{0.32\textwidth}
\vspace{0pt}
\centering
\scriptsize
\setlength{\tabcolsep}{2pt}

\resizebox{\linewidth}{!}{%
\begin{tabular}{lc}
\toprule
Read-time substitution
& $\mathcal{J}$\&$\mathcal{F}$ \\
\midrule
Correct memory
& $0.926$ \\

Wrong spatial memory
& $0.182$ \\

Wrong object pointer
& $0.927$ \\

Wrong spatial memory and pointer
& $0.183$ \\

Identity-free prototype
& $0.126$ \\

Zero memory
& $0.173$ \\
\bottomrule
\end{tabular}%
}

\captionof{table}{
Read-time memory substitution on DAVIS 2017 over ten sequences.
}
\label{tab-app-sam2-davis}

\end{minipage}

\end{table*}

\section{WorldMem Content Dependence}
\label{app-worldmem}

WorldMem \cite{xiao2025worldmem} tests a native memory system trained jointly with a generative world model. We use the $1000$ evaluation cases reported in the main paper.

Table~\ref{tab-app-worldmem} reports the exact aggregate results.


A wrong memory from another place in the same trajectory retains $94.1\%$ of the reference-relative benefit. A donor from a disjoint trajectory and different biome retains $43.7\%$. WorldMem therefore shows graded dependence on broader memory context. Exact place identity adds little beyond the same-trajectory control, while trajectory and scene agreement remain important.

WorldMem does not expose a true memory-off condition in this evaluation. We therefore use zero content as the content-free reference, and its performance defines $S_0$. Recovery measures content dependence relative to this active zero-content reference.

These controls change several forms of context at once. We therefore interpret their ordering as graded trajectory and scene dependence rather than as an additive decomposition of semantic factors.

\section{Additional SAM 2 Analyses}
\label{app-sam2}

SAM 2 \cite{sam2} provides the high-specificity comparison. We report the full DAVIS result, verify the intervention, and then analyze reappearance on MOSEv2.

\subsection{DAVIS Visible-Frame Substitution}
\label{app-sam2-davis}

Spatial memory is strongly content dependent on the evaluated DAVIS frames. We evaluate ten DAVIS 2017 sequences \cite{davis}. Table~\ref{tab-app-sam2-davis} reports the full channel-specific substitution result.

Replacing spatial memory removes most of the segmentation performance, while replacing the object pointer has almost no effect on these generally visible target frames. The spatial collapse is clear in eight of ten sequences. In the remaining two, current-frame evidence is sufficient to preserve the target. The result shows strong content dependence in the spatial memory under this setting.

\subsection{SAM 2 Intervention Verification}
\label{app-sam2-verification}

The small DAVIS pointer effect could arise if the pointer pathway were inactive or if the intervention failed to alter the computation. We test this directly on one DAVIS sequence.

The pointer intervention changes the internal representation even when the visible-frame prediction remains stable, as shown in Table~\ref{tab-app-sam2-verification}.

\begin{table}[t]
\footnotesize
\centering
\begin{tabular}{lc}
\toprule
Diagnostic & Value \\
\midrule
Object pointers used in encoder & True \\
Correct pointer IoU & $0.911$ \\
Wrong valid pointer IoU & $0.911$ \\
Zero pointer IoU & $0.909$ \\
Large random pointer IoU & $0.007$ \\
\midrule
Wrong spatial, memory-input change & $1.25$ \\
Wrong spatial, pre-decoder change & $1.01$ \\
Wrong spatial IoU & $0.000$ \\
\midrule
Wrong pointer, memory-input change & $0.023$ \\
Wrong pointer, pre-decoder change & $0.125$ \\
Wrong pointer IoU & $0.911$ \\
\bottomrule
\end{tabular}
\caption{SAM 2 intervention diagnostics on one DAVIS sequence using IoU.}
\label{tab-app-sam2-verification}
\end{table}

\begin{table}[t]
\footnotesize
\centering
\setlength{\tabcolsep}{4pt}
\renewcommand{\arraystretch}{1.05}
\begin{tabular}{llc}
\toprule
Condition & Metric & Estimate [95\% CI] \\
\midrule

Correct
& $\mathcal{J}$\&$\mathcal{F}$ & $0.459\ [0.371,0.548]$ \\
& Rec@1  & $0.435\ [0.352,0.522]$ \\
& Rec@5  & $0.520\ [0.442,0.598]$ \\
& Rec@10 & $0.551\ [0.476,0.625]$ \\
\addlinespace[2pt]

Wrong spatial
& $\mathcal{J}$\&$\mathcal{F}$ & $0.000\ [0.000,0.000]$ \\
& Rec@1  & $0.000\ [0.000,0.000]$ \\
& Rec@5  & $0.321\ [0.261,0.383]$ \\
& Rec@10 & $0.389\ [0.320,0.459]$ \\
\addlinespace[2pt]

Wrong pointer
& $\mathcal{J}$\&$\mathcal{F}$ & $0.385\ [0.299,0.474]$ \\
& Rec@1  & $0.360\ [0.277,0.443]$ \\
& Rec@5  & $0.480\ [0.401,0.557]$ \\
& Rec@10 & $0.521\ [0.445,0.598]$ \\
\addlinespace[2pt]

Wrong both
& $\mathcal{J}$\&$\mathcal{F}$ & $0.000\ [0.000,0.000]$ \\
& Rec@1  & $0.000\ [0.000,0.000]$ \\
& Rec@5  & $0.321\ [0.260,0.384]$ \\
& Rec@10 & $0.389\ [0.320,0.458]$ \\
\addlinespace[2pt]

Identity-free
& $\mathcal{J}$\&$\mathcal{F}$ & $0.003\ [0.000,0.007]$ \\
& Rec@1  & $0.001\ [0.000,0.004]$ \\
& Rec@5  & $0.342\ [0.281,0.404]$ \\
& Rec@10 & $0.413\ [0.343,0.481]$ \\
\addlinespace[2pt]

Zero
& $\mathcal{J}$\&$\mathcal{F}$ & $0.036\ [0.012,0.066]$ \\
& Rec@1  & $0.035\ [0.009,0.068]$ \\
& Rec@5  & $0.361\ [0.297,0.426]$ \\
& Rec@10 & $0.434\ [0.362,0.505]$ \\

\bottomrule
\end{tabular}
\caption{MOSEv2 point estimates with $95\%$ bootstrap confidence intervals from $10{,}000$ resamples.}
\label{tab-app-sam2-bootstrap}
\end{table}

A valid wrong pointer changes the internal representation even though the visible-frame output remains stable. A very large random pointer can collapse the prediction, showing that the pointer pathway can affect the output. Spatial-memory substitution produces a much larger internal change and collapses the mask on this diagnostic sequence. The weak DAVIS pointer effect is therefore not explained by an inactive intervention.

\subsection{MOSEv2 Reappearance}
\label{app-sam2-mose}

We evaluate $100$ MOSEv2 reappearance events \cite{mose,mosev2}. Each event contains a target that is visible, absent for at least five frames, and then visible again. The intervention is applied only at the first reappearance read. The preceding memory trajectory remains unchanged.

Correct spatial memory is critical at the return frame. Table~\ref{tab-app-sam2-mose} reports the exact values underlying the main-paper figure.


Wrong spatial memory and the identity-free prototype reduce return-frame performance to essentially zero. The object pointer also contributes at reappearance, but its effect is smaller. Later performance partially recovers because only the first reappearance read is changed. Subsequent predictions return to the intact memory state. Target transfer remains near zero, so spatial-memory corruption usually causes the target to be lost rather than transferred to the donor object.

\subsection{Bootstrap Uncertainty on MOSEv2}
\label{app-sam2-bootstrap}

We add uncertainty estimates to test whether the reappearance result is driven by a small subset of events. All $100$ events in this evaluation come from distinct videos. We use $10{,}000$ bootstrap resamples over events, which is equivalent to resampling videos for this set.

The strong spatial-memory effect remains under bootstrap resampling, as shown in Table~\ref{tab-app-sam2-bootstrap}.

Wrong-spatial return-frame performance remains at zero across the bootstrap samples. The spatial-memory dependence is therefore not driven by a small number of events. Wrong-pointer performance remains substantially higher, preserving the channel-specific difference.

\subsection{Object Pointer on Re-Detected Events}
\label{app-sam2-redetected}

The aggregate reappearance score includes cases where SAM 2 does not reacquire the target even with correct memory. We therefore examine the $49$ events where the object is re-detected under the correct-memory condition. On this subset, replacing the object pointer reduces $\mathcal{J}\&\mathcal{F}$ from $0.910$ to $0.757$, an absolute drop of $0.153$. The object pointer is therefore load bearing when the target must be reacquired, even though it has little effect on the generally visible DAVIS frames.

\subsection{Dependence Across Disappearance Durations}
\label{app-sam2-duration}

We test whether the spatial-memory result is limited to one disappearance duration. We divide the $100$ MOSEv2 events into three groups based on the number of frames for which the target is absent. Spatial-memory substitution reduces return-frame performance to approximately zero in every duration group, as shown in Figure~\ref{fig-sam2-duration}(a).



Correct-memory performance is lower in the longer-gap groups, indicating that these groups are harder on average. Spatial-memory dependence remains strong across the full range. The object pointer contributes in all three groups, but its absolute effect is not monotonic with disappearance duration.

Later recovery shows the same difficulty pattern in Figure~\ref{fig-sam2-duration}(b). Later recovery is lower in the longer-gap groups under both conditions. We therefore interpret disappearance duration primarily as a measure of task difficulty. Spatial-memory dependence remains present across all three groups.


\end{document}